# Deep Ordinal Ranking for Multi-Category Diagnosis of Alzheimer's Disease using Hippocampal MRI data

*Hongming Li, Mohamad Habes, Yong Fan*

*and for the Alzheimer's Disease Neuroimaging Initiative*\*

Section for Biomedical Image Analysis (SBIA), Center for Biomedical Image Computing and Analytics (CBICA), Department of Radiology, Perelman School of Medicine, University of Pennsylvania, Philadelphia, PA, 19104, USA

---

\*Data used in preparation of this article were obtained from the Alzheimer's Disease Neuroimaging Initiative (ADNI) database (adni.loni.usc.edu). As such, the investigators within the ADNI contributed to the design and implementation of ADNI and/or provided data but did not participate in analysis or writing of this report. A complete listing of ADNI investigators can be found at: http://adni.loni.usc.edu/wp-content/uploads/how_to_apply/ADNI_Acknowledgement_List.pdf

## Abstract

Increasing effort in neuroimaging has been dedicated to early diagnosis of Alzheimer's disease (AD) based on structural magnetic resonance imaging (MRI) data. Most existing studies have been focusing on binary classification problems, e.g., distinguishing AD patients from normal control (NC) elderly or mild cognitive impairment (MCI) individuals from NC elderly. However, identifying individuals with AD and MCI, especially MCI individuals who will convert to AD (progressive MCI, pMCI), in a single setting, is needed to early diagnose AD. In this paper, we propose a data-driven, deep ordinal ranking model for distinguishing NC, stable MCI (sMCI), pMCI, and AD at an individual subject level, taking into account the inherent ordinal severity of brain degeneration caused by normal aging, MCI, and AD, rather than formulating the classification as a traditional multi-category classification problem. The proposed deep ordinal ranking model focuses on the hippocampal morphology of individuals and learns informative and discriminative features automatically. We experimented with baseline MRI scans of 1776 subjects obtained from the Alzheimer's Disease Neuroimaging Initiative (ADNI) 1, ADNI GO, and ADNI 2. Our deep learning model was trained based on the ADNI 1 data and validated on the independent cohort of ADNI GO and ADNI 2. Our results indicate that the proposed method can achieve better performance than traditional multi-category classification techniques using shape and radiomics features from structural MRI data. Our method might accelerate the development of personalized AD diagnostic systems with targeted interventions.

## Introduction

Alzheimer's disease (AD) is the most prevalent neurodegenerative disorder. As a major public health issue, this disease results in tremendous neurologic disability, emotional suffering, and financial difficulty for patients, their families, and the society at large. Mild cognitive impairment (MCI) as a prodromal stage to AD, characterized by gradual neurodegeneration, is considered at a significantly higher risk to develop AD, with a conversion rate of 10-15% per year (Grundman *et al.*, 2004). Although clinical criteria for MCI and early AD have been developed to formalize assessment of the gradual progression of cognitive and other symptoms in early AD, currently it is difficult to predict which individuals who meet criteria for MCI will ultimately progress to AD. Neuroimaging has been playing an increasingly important role for clinical AD diagnostics. As the search for effective therapies to slow the progression of AD intensifies, there is a need for better diagnostic and prognostic tools to identify individuals at high risk to progress to AD.

To aid AD diagnosis and distinguish MCI patients with higher risk of conversion to AD (progressive MCI, pMCI) from stable MCI individuals (sMCI), machine learning techniques have been proposed to build classifiers upon imaging data and clinical measures (Davatzikos *et al.*, 2008; Fan *et al.*, 2008a; Fan *et al.*, 2008b; Misra *et al.*, 2009; Desikan *et al.*, 2010; Filipovych *et al.*, 2011; Moradi *et al.*, 2015; de Vos *et al.*, 2016; Hu *et al.*, 2016; Rathore *et al.*, 2017), and identified prominent structural differences between pMCI and sMCI subjects at medial temporal lobe (MTL), including regions such as hippocampus and entorhinal cortex.

Most existing classification studies of AD have been focusing on two-category classification problems, e.g., distinguishing AD patients from cognitively normal control (NC) elderly, MCI from NC, or pMCI from sMCI. However, the early diagnosis of AD is essentially a multi-category classification problem, i.e., we need to identify individuals with AD, pMCI, and sMCI in a single setting. The multi-category classification problem associated with early diagnosis of AD can be solved in a typical multi-category classification framework, using strategies of one-against-one or one-against-the-rest (Chih-Wei and Chih-Jen, 2002). However, such typical multi-category classification methods may overlook the ordinal information of the brain degeneration associated with rendered by normal aging, MCI and AD (Fan, 2011). Roughly speaking, brain changes rendered by normal aging, sMCI, pMCI, and AD come with an increased severity of the brain degeneration that is ordered, but the brain degeneration severity distances between the different stages is not known. The inter-subject variability might obliterate relatively small differences between NC and sMCI, between sMCI and pMCI, as well as between pMCI and AD, which makes it a difficult task for discriminating different stages of AD progression. Since no proper metric distance can be defined for the ordinal brain degeneration severity, metric regression methods might be not good for the problem too (Winship and Mare, 1984).

The hippocampus is one of the first brain structures affected by AD and undergoes severe structural changes (Braak and Braak, 1991). The structural variation between the hippocampus of AD patients and healthy subjects has been studied intensively (Qiu *et al.*, 2008; Teng *et al.*, 2015; Tsao *et al.*, 2017) (Li *et al.*, 2007; Chupin *et al.*, 2009; Gerardin *et al.*, 2009; Costafreda *et al.*, 2011; Devanand *et al.*, 2012; Ben Ahmed *et al.*, 2015; de Vos *et al.*, 2016; Hu *et al.*, 2016; Sorensen *et al.*, 2016; Aderghal *et al.*, 2017; Tsao *et al.*, 2017). Several studies have specifically focused on the hippocampus for early diagnosis of AD and build predictive models upon anatomical features including volume and shape based measures, and image intensity texture features (Chupin *et al.*, 2009; Devanand *et al.*, 2012; Ben Ahmed *et al.*, 2015; de Vos *et al.*, 2016; Hu *et al.*, 2016; Aderghal *et al.*, 2017; Tsao *et al.*, 2017). Particularly, promising performance of hippocampus shape (Li *et al.*, 2007; Gerardin *et al.*, 2009; Costafreda *et al.*, 2011), texture features (Sorensen *et al.*, 2016), and 2D convolutional neural networks (CNNs) based features (Aderghal *et al.*, 2017) has been demonstrated in AD prediction. However, most of the hippocampus focused pattern classification studies have been relying on the two-category classification techniques.

To achieve early prediction of AD based on the hippocampal MRI data, we develop an ordinal ranking based deep learning method, referred to as Deep Ordinal Ranking hereafter, to simultaneously learn reproducible and discriminative features from the hippocampal MRI data and classify AD, pMCI, sMCI, and NC subjects, by making the best of inherent ordinal severity of the brain degeneration at AD's different stages. Since deep convolutional neural networks (CNNs) based feature learning is potentially able to capture complex relationship between imaging data and the ordinal severity of the brain degeneration of AD, we adopt the CNNs to learn informative features from structural MRI data by optimizing a multi-output logistic regression model which encodes the ranking information of different stages of AD. We have evaluated the proposed method based on a large cohort of subjects from Alzheimer's Disease Neuroimaging Initiative (ADNI), including ADNI 1, ADNI GO, and ADNI 2. We compared the Deep Ordinal Ranking method with the state-of-the-art methods with multi-category classification capability. Experimental results have demonstrated that the proposed method could achieve improved prediction performance.

## Materials and Methods

*Image dataset*

The data used in this study were obtained from the ADNI cohort (http://adni.loni.usc.edu), consisting of baseline MRI scans of 1776 subjects from ADNI 1, ADNI Go and ADNI 2. The ADNI was launched in 2003 as a public-private partnership, led by Principal Investigator Michael W. Weiner, MD. The primary goal of ADNI has been to test whether serial MRI, positron emission tomography (PET), other biological markers, and clinical and neuropsychological assessment can be combined to measure the progression of MCI and early AD. For up-to-date information, see www.adni-info.org. We used MRI data (total n=817 scans, with 228 NC, 236 sMCI, 161 pMCI, and 192 AD patients) from ADNI 1 to train the proposed classification model. Then we validated the Deep Ordinal Ranking method with independent data (total n=959 scans with 311 NC, 395 sMCI, 94 pMCI and 158 AD patients) from the cohorts ADNI GO and ADNI 2. MCI subjects that converted to AD from 0.5 to 3 years from the baseline scan were defined as pMCI, otherwise they were considered as sMCI. The characteristics of the cohorts included in this study are summarized in table 1.

Table 1. Demographic and clinical diagnosis information of the subjects included in this study.

| ADNI | | NC | sMCI | pMCI | AD |
|---|---|---|---|---|---|
| 1 | Age | 75.97±5.02 | 75.03±7.67 | 74.58±7.00 | 75.34±7.45 |
| | Sex (M/F) | 118/110 | 156/80 | 100/61 | 101/91 |
| | MMSE | 29.11±1.00 | 27.31±1.78 | 26.63±1.69 | 23.31±2.04 |
| GO & 2 | Age | 72.98±6.09 | 71.44±7.56 | 72.60±7.02 | 74.85±8.09 |
| | Sex (M/F) | 142/169 | 213/182 | 53/41 | 91/67 |
| | MMSE | 29.00±1.25 | 28.24±1.60 | 27.23±1.84 | 23.1±2.07 |

NC: cognitively normal control; MCI: mild cognitive impairment; sMCI: stable MCI; pMCI: progressive MCI; AD: Alzheimer's diease. MCI subjects that converted to AD from 0.5 to 3 years from the baseline scan were defined as pMCI, otherwise defined as sMCI.

*Hippocampus extraction*

T1 MRI scans of all the subjects were registered to the MNI space using affine registration and resampled with a spatial resolution of $1 \times 1 \times 1$ mm$^3$. Left and right hippocampus regions were then segmented from the T1 images for each subject using the local label learning (LLL) (Hao *et al.*, 2014) algorithm with 100 hippocampus atlases obtained from a preliminary release of the EADC-ADNI harmonized segmentation protocol project (www.hippocampal-protocol.net) (Boccardi *et al.*, 2015). A 3D bounding box of size $29 \times 21 \times 55$ was adopted to extract hippocampus regions from the T1 image using the segmentation label of left and right hippocampus for each subject. These hippocampus regions, referred to as hippocampal MRI images hereafter, were used as the input to the proposed deep ordinal ranking model.

*Ordinal ranking*

To make the best of the ordinal severity of the brain degeneration rendered by normal aging, MCI, and AD, we propose an ordinal ranking method within an ordinal regression framework by transferring the ordinal ranking problem into a set of binary "larger than" problems (Fan, 2011). Particularly, NC, sMCI, pMCI and AD are labeled using an ordinal order $y \in \{1,2,3,4\}$, corresponding to their severity of the brain degeneration. Three binary "larger than" problems associated with the ordinal ranking problem are the brain degeneration "larger than normal aging?" ($y > 1$), "larger than sMCI?" ($y > 2$), and "larger than pMCI"($y > 3$). The binary "larger than" problems are solved separately and then the binary codes obtained are fused to obtain the final multi-category classification label.

Given training data $\{(x_i, y_i), i = 1,2,...,n\}$, where $x_i$ represents the feature vector for subject $i$ and $y_i \in \{1,2,3,4\}$ represents its associated category label, the 4-category label could be transformed into a binary label for each binary "larger than" problem. For the $k$-th binary problem ($y > k$), its positively labeled training dataset $X_k^+$ and negatively labeled training dataset $X_k^-$ could be constructed as

$$X_k^+ = \{(x_i, 1) | y_i > k\}, X_k^- = \{(x_i, 0) | y_i \leq k\}. \tag{1}$$

Based on the training dataset, a binary classifier $f_k$ could be trained using any pattern classification techniques, such as support vector machine (SVM) (Cortes and Vapnik, 1995) and random forests (RF) (Tin Kam, 1998). Once all the three binary classifiers are obtained, the ordinal ranking rule is constructed as

$$r(x) = 1 + \sum_{k=1}^{3} [\![f_k(x) > 0]\!], \tag{2}$$

where $[\![\cdot]\!]$ is 1 if the inner condition is true and 0 otherwise.

*Deep ordinal ranking*

Given the imaging data of hippocampus of each subject, different kinds of feature representations could be extracted, such as shape representation and radiomic characterization of image texture measures within the hippocampus regions (Rathore *et al.*, 2017). Although these representations have been investigated and achieved promising performance, as hand-crafted features they might be not optimal and less discriminative for the AD diagnosis.

The success of deep learning techniques in pattern recognition (Goodfellow *et al.*, 2016) in recent years have witnessed promising performance in learning imaging features for a variety of pattern recognition tasks (Gulshan *et al.*, 2016; Nie *et al.*, 2016; Esteva *et al.*, 2017). In these studies, convolutional neural networks (CNNs) are widely adopted to learn informative imaging features by optimizing a pattern recognition cost function. Ordinal regression based on CNNs has also been adopted for age estimation, and achieved better performance than state-of-the-art alternative techniques (Niu *et al.*, 2016). Therefore, we propose a deep ordinal model for AD diagnosis based on CNNs to learn informative and discriminative feature representation of the hippocampus and the mapping between the deep features and ordinal ranking in a data-driven way simultaneously.

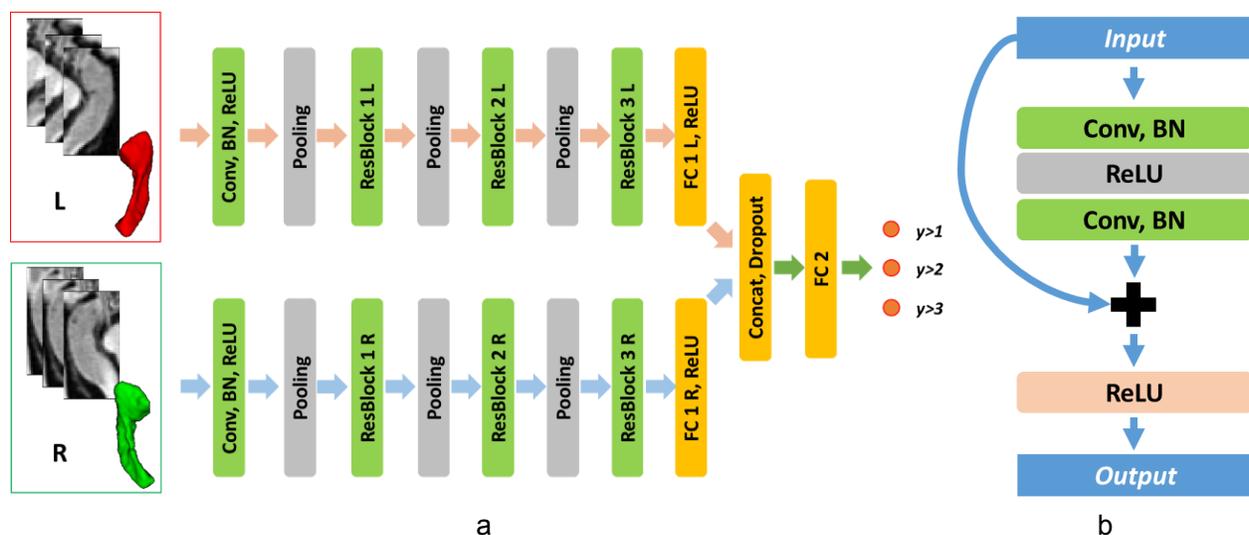

Fig. 1. Deep ordinal ranking model for data-driven hippocampus-based early AD diagnosis. (a) schematic architecture of the deep network, (b) schematic residual block. L: left hippocampus; R: right hippocampus.

The network architecture of the proposed deep learning model is illustrated in Fig. 1a. The Deep Ordinal Ranking model contains one convolutional layer (Conv), followed by three residual blocks (ResBlock), one fully connected layer (FC), and an output layer for the ordinal ranking (FC2). Rectified linear units (ReLU) is used as a nonlinear activation function for the convolutional and fully connected layers, batch normalization (BN) is adopted to accelerate deep network training (Ioffe and Szegedy, 2015), and max pooling layers are adopted to obtain features at multiple scales. The residual network structure, as illustrated in Fig. 1b, has been adopted widely since its invention (He *et al.*, 2016) and achieved promising performance in many challenging pattern recognition tasks. Several studies have also demonstrated that the residual connection would accelerate the convergence and improve the performance of the CNNs (Szegedy *et al.*, 2017). The left and right hippocampus regions are adopted as two-stream inputs to the deep model, which are gradually convolved by multiple 3D kernels within the subsequent Conv layer and ResBlock layers. The high-level feature representations of each hippocampus region are then flatten and connected to the FC layers, whose output are concatenated and fed into the output layer.

To learn imaging features informative for the ordinal ranking with binary "larger than" classification problems, we formulate the ordinal ranking as a multi-label classification problem. In our study, the four-category label of each subject $y_i \in \{1,2,3,4\}$ is transformed into a 3-bit binary label encoding its status corresponding to the 3 binary "larger than" problems, i.e., the brain degeneration "larger than normal aging?", "larger than sMCI?", and "larger than pMCI?". For example, one AD patient will be labeled as [1,1,1] in the deep ordinal ranking setting while labeled as [0,0,0,1] in the regular four-category classification setting. The output layer has three nodes corresponding to the three binary "larger than" problems in the Deep Ordinal Ranking model. Sigmoid cross entropy loss is adopted to optimize the deep learning model.

*Data augmentation*

To boost the deep learning model's performance and robustness to image alignment and hippocampus segmentation errors, data augmentation is adopted to generate more training data (Goodfellow *et al.*, 2016). Particularly, augmented image data were generated using image translation and non-rigid deformable image registration techniques. In particular, each hippocampus image along with its corresponding hippocampus masks in the training dataset was translated by 2 voxels along 26 directions of 3D image space separately, yielding augmented images that account for translation invariance for training the deep learning model. A non-rigid deformable image registration method, namely ANTs (Avants *et al.*, 2011), was adopted with its default parameter setting to register one hippocampal MRI image, referred to as moving image, to another of the same side (left to left and right to right) within the same disease category (NC to NC, sMCI to sMCI, pMCI to pMCI, and AD to AD), and the resulting deformation field was used to deform the moving hippocampus image and its hippocampus label to generate deformed hippocampus image and label. In total, 21242 spatial translated images, and 84824 non-rigid registered images were generated as the augmented dataset for training the deep learning model.

## Validation and comparisons

We evaluated the proposed method and compared it with state-of-the-art alternative methods based on the same training and validation datasets.

*State-of-the-art alternative methods under comparison*

We compared the Deep Ordinal Ranking method with the state-of-the-art feature extraction methods for hippocampal MRI images with the regular multi-category classification and ordinal ranking settings. The feature extraction methods for the hippocampal MRI images include shape characterization, tools of radiomics for extracting texture features (Griethuysen *et al.*, 2017), and deep CNNs. Details of these classification schemes are as following.

- Hippocampal feature extraction.
  - Shape characterization: 11 shape related features are extracted from the segmentation label of left and right hippocampus respectively, including volume, maximum 3D diameter, maximum 2D diameter (column, row, and slice), surface area, surface volume ratio, flatness, sphericity, elongation, and spherical disproportion (Griethuysen *et al.*, 2017).
  - Radiomics: image texture features are extracted from the hippocampal images and their counterparts after wavelet decomposition, including the first order features, gray level co-occurrence matrix (GLCM) features, gray level size zone matrix (GLSZM) features, and gray level run length matrix (GLRLM) features, and there are 711 textures in total for each hippocampus region. The shape and texture features are calculated using the pyradiomics packages (http://pyradiomics.readthedocs.io) (Griethuysen *et al.*, 2017).
  - Deep representation: informative and discriminative features are automatically learned during the training procedure of forward convolution and back-propagation of deep CNNs.
- Classifier construction.
  - Shallow classifier: Random forests (RF) (Tin Kam, 1998) is adopted to construct classifier using shape and radiomics representation respectively. Its inherent feature selection and decision ensemble techniques lead to robust classification and better generalization. Moreover, RF can handle multi-category classification naturally.
  - Deep classifier: CNNs with the architecture shown in Fig. 1 is adopted for the prediction tasks based on the learned features in the data-driven way.

- Classification strategy.
    - Regular multi-category classification: the early diagnosis of AD is formulated as a 4-category classification problem, RF using shape representation, RF using radiomics representation, and CNNs with 4-category output are evaluated in this study. For the CNNs, the network architecture is the same as that in Fig.1 except that the output layer is replaced with 4-node output layer for regular multi-category classification.
    - Ordinal ranking classification: for the shallow classifier, 3 "larger than" binary classifiers are constructed using RF based on shape and radiomics representation respectively. For the deep classifier, the proposed deep ordinal ranking model as illustrated in Fig.1 is adopted. Note that the same network architecture and parameter configuration is adopted for deep classifier under both regular multi-category and ordinal ranking setting except the differences of the output layer.

In addition to the 4-category classification, we have also performed a binary classification to distinguish AD patients from NC elderly based on shape, radiomics, and deep CNNs representation as baseline experiments, in order to investigate the discriminative power of hippocampal representation for AD diagnosis.

The performance of the classification is evaluated with the following metrics: (1) normalized confusion matrix, (2) adjusted classification accuracy, (3) receiver operating characteristic curve (ROC), and area under ROC (AUC). A normalized confusion matrix illustrates not only the sensitivity and specificity for the multi-category classification results, but also the pattern of misclassification reflecting the severity of different stages of AD disease. Adjusted classification accuracy is calculated as the mean sensitivity value of the 4 categories, which takes the imbalance of sample sizes of different categories into consideration. For the binary AD versus NC prediction, ROC and AUC are adopted for the evaluation.

*Experimental settings*

The deep learning model's network architecture is illustrated by Fig. 1, with 1 Conv layer, 3 ResBlocks, 1 FC layer, and an output layer. In particular, the Conv layer contains 64 kernels, while the ResBlock 1, 2, and 3 contains 64, 128, and 128 kernels respectively. The kernel size for all the kernels is $3 \times 3 \times 3$. A stride of 2 and kernel size of 2 is used for the max pooling layer. The fully connected layer FC1 contains 256 nodes, which extract a 256-dimensional features for left and right hippocampus respectively. The two 256-dimensional feature vector is concatenated and fed to FC2 with 3 output nodes for the deep ordinal ranking model (4 output nodes for the deep multi-category classification model). A dropout operation with a ratio of 0.5 is applied before the features fed into the last FC layer.

The deep learning model was optimized using stochastic gradient descent (SGD) algorithm (Boyd and Vandenberghe, 2004), the momentum was set to 0.9, and the base learning rate was set to $5 \times 10^{-5}$. The learning rate was updated using a stepwise policy, which drops the learning rate by a factor of 0.1 after every 40000 steps. The maximum iteration of the training procedure was set to 120000. Batch size of 32 was adopted to update weights in the model. The deep learning models was implemented using Caffe (Jia *et al.*, 2014), and trained on a Nividia Titan X (Pascal) graphics processing unit (GPU).

For the RF based on shape representation and radiomics representation, 1000 decision trees were adopted, and the minimum leaf size of the tree was set to 5. Sample weight for each training image was set to the ratio between total number of training images and the number of images within the same category, and the training images were sampled with replacement during the training procedure. The built-in RF implementation TreeBagger in Matlab (R2013a) was adopted to train the model, and default values were used for other parameters.

# Results

The distributions of hippocampus volumes of different groups are illustrated in Fig. 2. These plots indicated that AD patients and NC elderly could be roughly separated based on their hippocampus volumes. However, the hippocampus volumes of MCI individuals scatter in-between the AD group and NC group, demonstrating the complexity of distinguishing between the 4 groups based on the hippocampus volume only. In fact, that might be impossible.

Two experiments were conducted to evaluate the performance of the Deep Ordinal Ranking model. We first performed a binary classification task for distinguishing the AD patients from the NC individuals using the shape representation, radiomics representation, and deep representation respectively with a two-fold purpose. On one hand, we would like to check the power of hippocampus based representation for the AD diagnosis, on the other hand, we would like to investigate if the deep representation learned based on the deep CNNs is more discriminative for the prediction task. We then performed the 4-category prediction based on the 3 kinds of hippocampus representation under regular multi-category classification and ordinal ranking setting, to investigate if improved prediction performance could be achieved by the Deep Ordinal Ranking model. It is worth noting that all the prediction models were trained using the ADNI I dataset, and validated using the ADNI Go & 2 dataset.

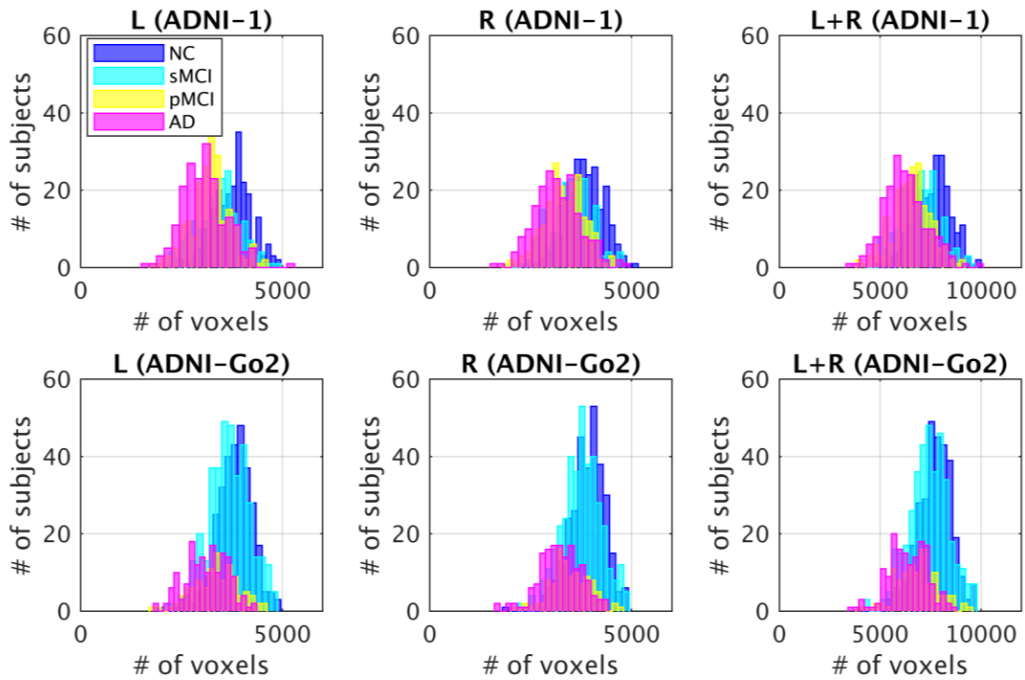

Fig. 2. Histograms of the hippocampus volume measures of all the subjects (top row: ADNI 1 and bottom row: ADNI GO & 2) used in this study. Volume measures of left (L) and right (R) hippocampi and their combination are shown from left to right. Each voxel has a spatial resolution of 1x1x1mm$^3$.

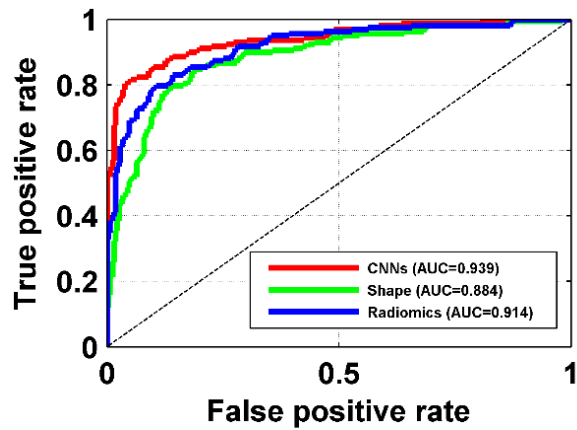

Fig. 3. ROC curves obtained based on different hippocampus representations for AD versus NC prediction.

Fig. 3 shows the ROC curve of the binary classification on the validation dataset using the shape, radiomics, and deep representations respectively. The AUC obtained by the deep representation was 0.939, while those obtained based on the shape representation and radiomic representation were 0.884 and 0.914 respectively. As demonstrated in Fig. 3, all the 3 hippocampus representations were quite powerful for distinguishing AD patients from NC subjects, indicating that hippocampus based representations could characterize the anatomical alternations caused by the disease. Moreover, the radiomics representation obtained better performance than the shape representation, and the deep representation obtained the best performance, indicating that intensity variations within hippocampus could provide more discriminative information, and features learned in the data-driven manner could capture the task related characteristics better than the conventional hand-crafted features.

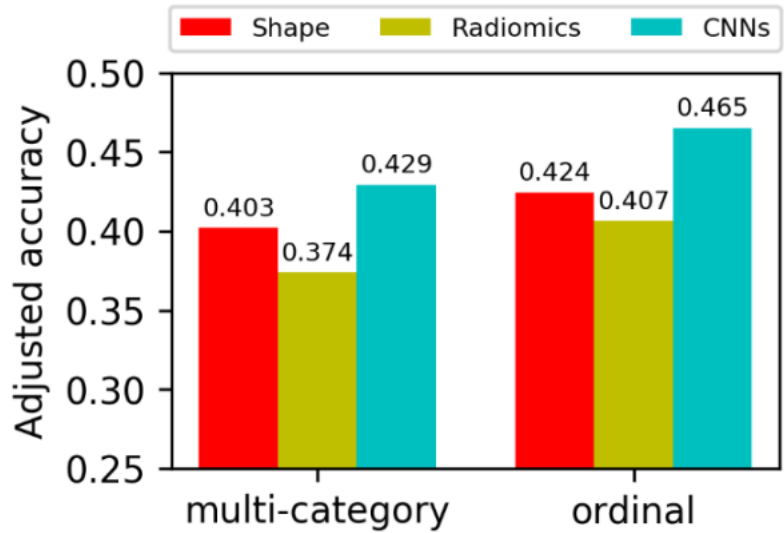

Fig. 4. Adjusted accuracy for the 4-category prediction under different classification setting.

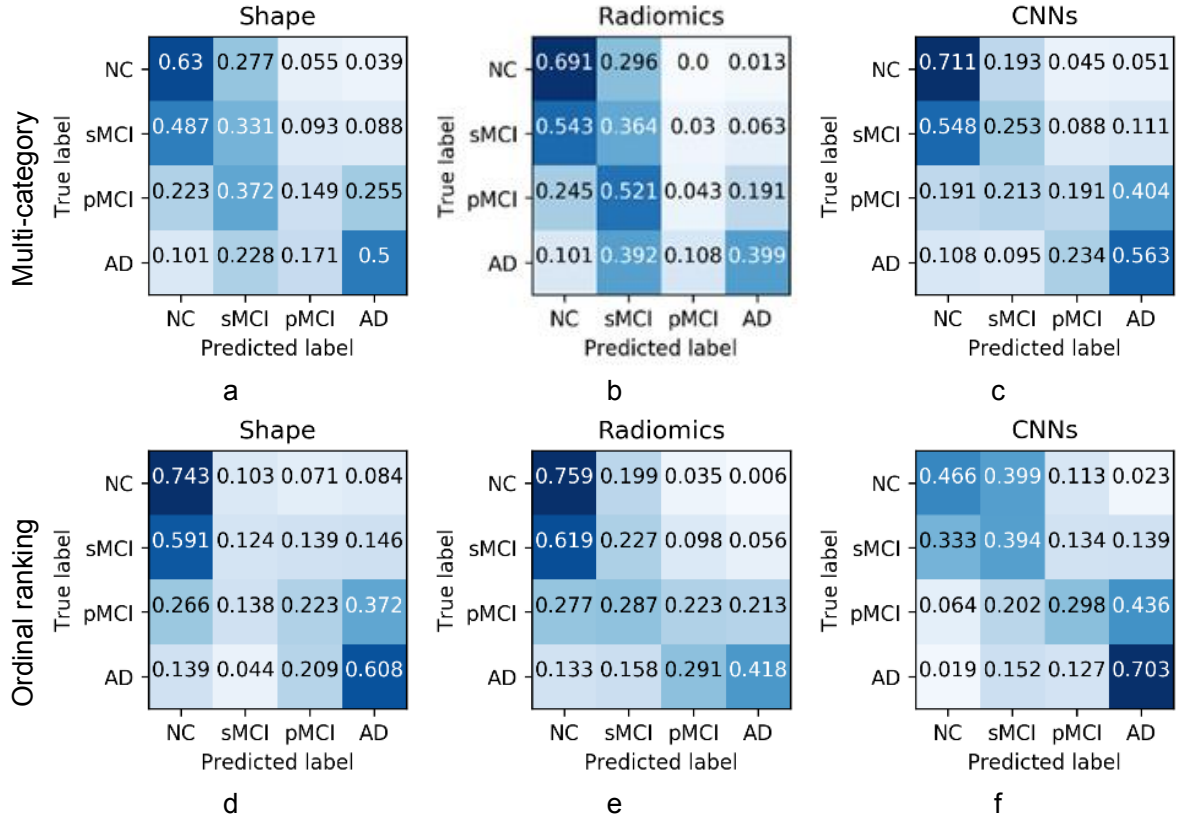

Fig. 5. Confusion matrices of 4-category prediction. (a-c) results obtained based on different hippocampus representations under regular multi-category classification setting, (d-f) corresponding results under ordinal ranking setting.

The results of 4-category prediction are illustrated in Fig. 4 and Fig. 5. Fig. 4 shows the adjusted accuracy for the prediction. It could be observed that the CNNs model obtained better performance than the RF method using shape and radiomics representations, and the performance under the ordinal ranking setting were generally better than their counterparts under the conventional multi-category classification setting. The best performance was obtained by the Deep Ordinal Ranking model, and the adjusted accuracy was 0.465. Fig. 5 illustrates the confusion matrices of all the 6 prediction models. Generally speaking, the AD group and NC group were separated pretty well by all the models, our Deep Ordinal Ranking model captured the progressive patterns of the AD better than other models, as the larger coefficients of the confusion matrix located at the nearby positions along the diagonal of the matrix, indicating that misclassified subjects were assigned to adjacent categories in the progression spectrum, instead of the distant categories.

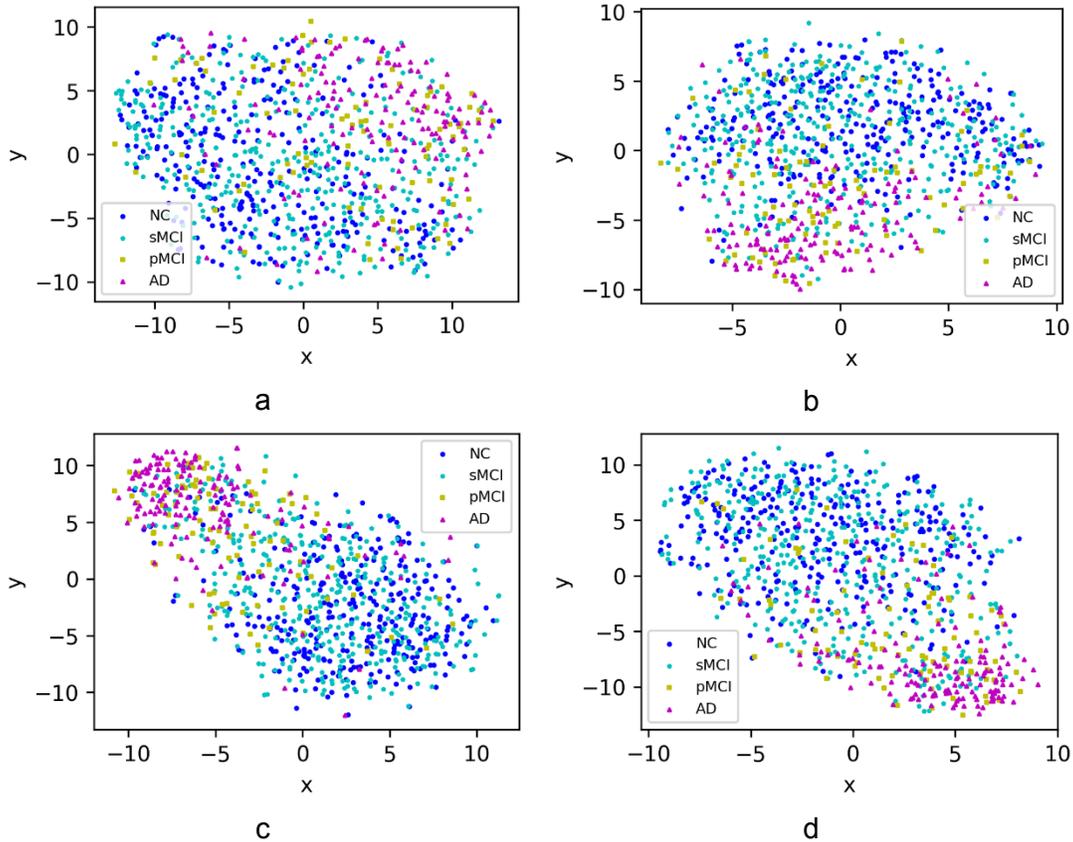

Fig. 6. t-SNE visualization of different hippocampus representation. (a) shape, (b) raidomics, (c) multi-category CNNs, and (d) Deep Ordinal Ranking CNNs.

To investigate how the different hippocampus representations contributed to the classification, we have projected the different hippocampus representations onto a 2D plane using the t-SNE algorithm (Maaten and Hinton, 2008), as shown in Fig. 6. The 4 subplots correspond to the shape representation, radiomics representation, deep representation learned by multi-category CNNs, and deep representation learned by our Deep Ordinal Ranking method. As illustrated in Fig. 6, for the shape and radiomics representations, the distribution of the sMCI and pMCI individuals were largely overlapped with those of AD and NC individuals, which limited the discriminative power of the corresponding prediction models. For the deep representations, a relative clear progressive pattern could be observed, where the AD and NC individuals distributed around top-left and bottom-right corners while the sMCI and pMCI individuals spanned in between. The visualization results also indicated that the learned representations were more informative and facilitated the subsequent prediction.

## Discussions

As one of the first brain structures affected by AD, the structural variation of hippocampus between AD patients and healthy subjects has been studied intensively. Although several studies have been proposed to extract different feature representations of hippocampus from structural MR imaging data for computer-aided AD diagnosis, most of them focus on shape related features or conventional hand-crafted radiomic features, which might not be discriminative for the diagnosis task. Moreover, these studies generally focus on binary classification instead of 4-category prediction covering all the stages of AD progression which is more clinically relevant, and do not take the intrinsic ordinal severity of different stages of AD into account. To this end, we develop a deep ordinal ranking framework to automatically extract hippocampus representation from MR images in a data-driven way and the mapping between them and the ordinal AD staging information simultaneously. In particular, we get the deep representations for left and right hippocampus respectively using deep CNNs which extract imaging features hierarchically. The regular 4-category labels are transformed into 3 ordinal labels which are used for optimizing the multi-output loss function to drive the whole deep learning model. We have trained and validated our Deep Ordinal Ranking model with baseline MRI scans of 1776 subjects from independent sub-cohorts of the ADNI dataset, and compared our method with state-of-the-art feature extraction and pattern classification techniques. Particularly, imaging data obtained from ADNI 1 were used as training dataset and those obtained from ADNI Go & 2 were used as independent validation dataset. The experimental results have demonstrated that the proposed method could help improve the multi-category diagnosis of AD. It would be straightforward to integrate multimodal imaging data and biological/clinical measures using the convolutional layers and fully connected layers.

Several quantitative measures about hippocampus have been explored to investigate their discriminative power to distinguish AD patients from NC individuals, from simple volume, to geometric shape measures, to intensity based imaging features such as texture features. Though statistical differences between patients and health individuals and promising classification performance have been reported based on these measures, which have also been evaluated as in our binary classification experiment, the discriminative power of these hand-crafted measures are still limited, especially when used for more complex classification tasks, such as the 4-category classification. As illustrated in Fig. 6 (a, b), they were unable to effectively separate the distributions of sMCI and pMCI individuals who are at an intermediate stage between AD and NC, and therefore their corresponding classification performance were hindered, as shown in Fig. 5(a, b, d, e). Unlike the hand-crafted feature extractors, the deep CNNs could extract relevant features tailored for specific classification tasks, and the extracted features could be optimized during the training procedure of the deep learning models. As illustrated in Fig. 6 (c, d), individuals of AD and NC were relatively better separated based on the deep learning features than based on the hand-crafted features, while the overall distributions of 4 categories showed a relatively clearer transition pattern from the top-left corner to the bottom-right corner. As expected, the deep learning features also promoted the classification performance as shown in Fig. 5 (c, f).

Instead of formulating the AD diagnosis as binary classification that accounts for 2 out 4 stages of AD progression, or as regular multi-category classification ignoring the progressive property of adjacent stages, we formulate the diagnosis task under an ordinal ranking framework. The ordinal ranking framework can naturally consider the severity degrees of brain degeneration along with the disease progression. Under regular multi-category classification setting, one subject might be misclassified into one arbitrary category. However, larger penalty would be introduced to the prediction model under the ordinal ranking setting if one pMCI individual is assigned to the NC instead of sMCI, as NC is more distant from pMCI on the ordinal list. This has also been demonstrated in Fig. 4 and 5. All the prediction models under ordinal ranking setting outperformed their multi-category counterparts, and the pattern of the prediction results followed the disease progression better, as shown in Fig. 5f in particular, most of the incorrectly assigned individuals were located at adjacent categories of their true category.

Although the proposed deep ordinal ranking model has achieved promising performance for AD diagnosis, further effort is needed in following aspects. First, the current study focused on the hippocampus in AD diagnosis, and obtained similar classification performance as those methods based on the whole brain information (Liu *et al.*, 2015b). It is expected to obtain improved classification performance to extract informative features from the whole brain MRI data. Second, hyper-parameters of the deep ordinal ranking model need further optimization, including network architecture, convolutional filter size, learning rate, batch size, number of filters per convolution layer, and so on (Goodfellow *et al.*, 2016). Currently, we set these parameters considering GPU memory. However, Bayesian optimization methods could be used to tune our models (Snoek *et al.*, 2012), and better performance could be obtained. Moreover, the definition of pMCI category might influence the performance of the diagnosis. Conversion to AD within 2 or 3 years are generally used for the identification of pMCI in the literature. However, other settings need to be considered.

## Conclusion

In this paper, we have presented a deep ordinal ranking model for classifying AD's different stages using structural imaging data focusing on the hippocampus, built on CNNs and ordinal ranking techniques. The comparison with the traditional multi-category classification methods based on the ADNI dataset has demonstrated that our method could achieve promising performance, indicating that the utilization of inherent ordinal severity of brain degeneration associated with AD's different stages could help achieve improved classification performance. Moreover, the deep learning features of the hippocampus also outperformed hand-crafted imaging features, i.e., shape and radiomics features. Benefiting from the flexible architecture of proposed deep model, the performance of our method might be further improved if multi-modality information is taken into account, e.g., PET imaging and CSF biomarkers (Liu *et al.*, 2015a). Besides classification, our proposed method is also a better fit for regression studies of AD associated clinical score estimation than simple metric regression, since most of the clinical score measures, e.g., mini mental state examination (MMSE), are not continuous variables.

## Acknowledgments

This work was supported in part by National Institutes of Health grants (Nos. EB022573, CA189523, MH107703, DA039215, and DA039002).

Data collection and sharing for this project was funded by the Alzheimer's Disease Neuroimaging Initiative (ADNI) (National Institutes of Health Grant U01 AG024904) and DOD ADNI (Department of Defense award number W81XWH-12-2-0012). ADNI is funded by the National Institute on Aging, the National Institute of Biomedical Imaging and Bioengineering, and through generous contributions from the following: AbbVie, Alzheimer's Association; Alzheimer's Drug Discovery Foundation; Araclon Biotech; BioClinica, Inc.; Biogen; Bristol-Myers Squibb Company; CereSpir, Inc.; Cogstate; Eisai Inc.; Elan Pharmaceuticals, Inc.; Eli Lilly and Company; EuroImmun; F. Hoffmann-La Roche Ltd and its affiliated company Genentech, Inc.; Fujirebio; GE Healthcare; IXICO Ltd.; Janssen Alzheimer Immunotherapy Research & Development, LLC.; Johnson & Johnson Pharmaceutical Research & Development LLC.; Lumosity; Lundbeck; Merck & Co., Inc.; Meso Scale Diagnostics, LLC.; NeuroRx Research; Neurotrack Technologies; Novartis Pharmaceuticals Corporation; Pfizer Inc.; Piramal Imaging; Servier; Takeda Pharmaceutical Company; and Transition Therapeutics. The Canadian Institutes of Health Research is providing funds to support ADNI clinical sites in Canada. Private sector contributions are facilitated by the Foundation for the National Institutes of Health (www.fnih.org). The grantee organization is the Northern California Institute for Research and Education, and the study is coordinated by the Alzheimer's Therapeutic Research Institute at the University of Southern California. ADNI data are disseminated by the Laboratory for Neuro Imaging at the University of Southern California.

# References

Aderghal K, Boissenin M, Benois-Pineau J, Catheline G, Afdel K. Classification of sMRI for AD Diagnosis with Convolutional Neuronal Networks: A Pilot 2-D+e Study on ADNI. In: Amsaleg L, Guðmundsson GÞ, Gurrin C, Jónsson BÞ, Satoh Si, editors. MultiMedia Modeling: 23rd International Conference, MMM 2017, Reykjavik, Iceland, January 4-6, 2017, Proceedings, Part I. Cham: Springer International Publishing; 2017. p. 690-701.

Avants BB, Tustison NJ, Song G, Cook PA, Klein A, Gee JC. A reproducible evaluation of ANTs similarity metric performance in brain image registration. Neuroimage 2011; 54(3): 2033-44.

Ben Ahmed O, Benois-Pineau J, Allard M, Ben Amar C, Catheline G. Classification of Alzheimer's disease subjects from MRI using hippocampal visual features. Multimedia Tools and Applications 2015; 74(4): 1249-66.

Boccardi M, Bocchetta M, Morency FC, Collins DL, Nishikawa M, Ganzola R, *et al.* Training labels for hippocampal segmentation based on the EADC-ADNI harmonized hippocampal protocol. Alzheimers Dement 2015; 11(2): 175-83.

Boyd S, Vandenberghe L. Convex optimization: Cambridge university press; 2004.

Braak H, Braak E. Neuropathological stageing of Alzheimer-related changes. Acta Neuropathol 1991; 82(4): 239-59.

Chih-Wei H, Chih-Jen L. A comparison of methods for multiclass support vector machines. IEEE Transactions on Neural Networks 2002; 13(2): 415-25.

Chupin M, Gerardin E, Cuingnet R, Boutet C, Lemieux L, Lehericy S, *et al.* Fully automatic hippocampus segmentation and classification in Alzheimer's disease and mild cognitive impairment applied on data from ADNI. Hippocampus 2009; 19(6): 579-87.

Cortes C, Vapnik V. Support-vector networks. Machine Learning 1995; 20(3): 273-97.

Costafreda SG, Dinov ID, Tu Z, Shi Y, Liu C-Y, Kloszewska I, *et al.* Automated hippocampal shape analysis predicts the onset of dementia in mild cognitive impairment. Neuroimage 2011; 56(1): 212-9.

Davatzikos C, Fan Y, Wu X, Shen D, Resnick SM. Detection of prodromal Alzheimer's disease via pattern classification of magnetic resonance imaging. Neurobiol Aging 2008; 29(4): 514-23.

de Vos F, Schouten TM, Hafkemeijer A, Dopper EG, van Swieten JC, de Rooij M, *et al.* Combining multiple anatomical MRI measures improves Alzheimer's disease classification. Hum Brain Mapp 2016; 37(5): 1920-9.

Desikan RS, Cabral HJ, Settecase F, Hess CP, Dillon WP, Glastonbury CM, *et al.* Automated MRI measures predict progression to Alzheimer's disease. Neurobiology of aging 2010; 31(8): 1364-74.

Devanand DP, Bansal R, Liu J, Hao X, Pradhaban G, Peterson BS. MRI hippocampal and entorhinal cortex mapping in predicting conversion to Alzheimer's disease. Neuroimage 2012; 60(3): 1622-9.

Esteva A, Kuprel B, Novoa RA, Ko J, Swetter SM, Blau HM, *et al.* Dermatologist-level classification of skin cancer with deep neural networks. Nature 2017; 542(7639): 115-8.

Fan Y. Ordinal Ranking for Detecting Mild Cognitive Impairment and Alzheimer's Disease Based on Multimodal Neuroimages and CSF Biomarkers. In: Liu T, Shen D, Ibanez L, Tao X, editors. Multimodal Brain Image Analysis: First International Workshop, MBIA 2011, Held in Conjunction with MICCAI 2011, Toronto, Canada, September 18, 2011 Proceedings. Berlin, Heidelberg: Springer Berlin Heidelberg; 2011. p. 44-51.

Fan Y, Batmanghelich N, Clark CM, Davatzikos C, Initia ADN. Spatial patterns of brain atrophy in MCI patients, identified via high-dimensional pattern classification, predict subsequent cognitive decline. Neuroimage 2008a; 39(4): 1731-43.

Fan Y, Resnick SM, Wu XY, Davatzikos C. Structural and functional biomarkers of prodromal Alzheimer's disease: A high-dimensional pattern classification study. Neuroimage 2008b; 41(2): 277-85.

Filipovych R, Davatzikos C, Alzheimer's Disease Neuroimaging I. Semi-supervised pattern classification of medical images: application to mild cognitive impairment (MCI). Neuroimage 2011; 55(3): 1109-19.

Gerardin E, Chetelat G, Chupin M, Cuingnet R, Desgranges B, Kim H-S, *et al.* Multidimensional classification of hippocampal shape features discriminates Alzheimer's disease and mild cognitive impairment from normal aging. Neuroimage 2009; 47(4): 1476-86.

Goodfellow I, Bengio Y, Courville A. Deep Learning: MIT Press; 2016.

Griethuysen JJv, Fedorov A, Parmar C, Hosny A, Aucoin N, Narayan V, *et al.* Computational Radiomics System to Decode the Radiographic Phenotype. Cancer Research 2017.

Grundman M, Petersen RC, Ferris SH, et al. Mild cognitive impairment can be distinguished from alzheimer disease and normal aging for clinical trials. Archives of Neurology 2004; 61(1): 59-66.

Gulshan V, Peng L, Coram M, et al. Development and validation of a deep learning algorithm for detection of diabetic retinopathy in retinal fundus photographs. JAMA 2016; 316(22): 2402-10.

Hao Y, Wang T, Zhang X, Duan Y, Yu C, Jiang T, *et al.* Local label learning (LLL) for subcortical structure segmentation: Application to hippocampus segmentation. Hum Brain Mapp 2014; 35(6): 2674-97.

He K, Zhang X, Ren S, Sun J. Deep Residual Learning for Image Recognition.  2016 IEEE Conference on Computer Vision and Pattern Recognition (CVPR); 2016 27-30 June 2016; 2016. p. 770-8.

Hu K, Wang Y, Chen K, Hou L, Zhang X. Multi-scale features extraction from baseline structure MRI for MCI patient classification and AD early diagnosis. Neurocomputing 2016; 175, Part A: 132-45.

Ioffe S, Szegedy C. Batch Normalization: Accelerating Deep Network Training by Reducing Internal Covariate Shift. In: Bach FR, Blei DM, editors. ICML: JMLR.org; 2015. p. 448-56.

Jia Y, Shelhamer E, Donahue J, Karayev S, Long J, Girshick R, *et al.* Caffe: Convolutional architecture for fast feature embedding.  Proceedings of the 22nd ACM international conference on Multimedia; 2014: ACM; 2014. p. 675-8.

Li S, Shi F, Pu F, Li X, Jiang T, Xie S, *et al.* Hippocampal shape analysis of Alzheimer disease based on machine learning methods. AJNR Am J Neuroradiol 2007; 28(7): 1339-45.

Liu L, Fu L, Zhang X, Zhang J, Zhang X, Xu B, *et al.* Combination of dynamic (11)C-PIB PET and structural MRI improves diagnosis of Alzheimer's disease. Psychiatry Res 2015a; 233(2): 131-40.

Liu S, Liu S, Cai W, Che H, Pujol S, Kikinis R, *et al.* Multimodal neuroimaging feature learning for multiclass diagnosis of Alzheimer's disease. IEEE transactions on bio-medical engineering 2015b; 62(4): 1132-40.

Maaten Lvd, Hinton G. Visualizing data using t-SNE. Journal of Machine Learning Research 2008; 9(Nov): 2579-605.

Misra C, Fan Y, Davatzikos C. Baseline and longitudinal patterns of brain atrophy in MCI patients, and their use in prediction of short-term conversion to AD: Results from ADNI. Neuroimage 2009; 44(4): 1415-22.

Moradi E, Pepe A, Gaser C, Huttunen H, Tohka J, Alzheimer's Disease Neuroimaging I. Machine learning framework for early MRI-based Alzheimer's conversion prediction in MCI subjects. Neuroimage 2015; 104: 398-412.

Nie D, Zhang H, Adeli E, Liu L, Shen D. 3D Deep Learning for Multi-modal Imaging-Guided Survival Time Prediction of Brain Tumor Patients. Medical image computing and computer-assisted intervention : MICCAI International Conference on Medical Image Computing and Computer-Assisted Intervention 2016; 9901: 212-20.

Niu Z, Zhou M, Wang L, Gao X, Hua G. Ordinal Regression with Multiple Output CNN for Age Estimation.  2016 IEEE Conference on Computer Vision and Pattern Recognition (CVPR); 2016 27-30 June 2016; 2016. p. 4920-8.

Qiu A, Younes L, Miller MI, Csernansky JG. Parallel transport in diffeomorphisms distinguishes the time-dependent pattern of hippocampal surface deformation due to healthy aging and the dementia of the Alzheimer's type. Neuroimage 2008; 40(1): 68-76.

Rathore S, Habes M, Iftikhar MA, Shacklett A, Davatzikos C. A review on neuroimaging-based classification studies and associated feature extraction methods for Alzheimer's disease and its prodromal stages. Neuroimage 2017; 155: 530-48.

Snoek J, Larochelle H, Adams RP. Practical Bayesian Optimization of Machine Learning Algorithms.  Advances in Neural Information Processing Systems; 2012. p. 1-9.

Sorensen L, Igel C, Liv Hansen N, Osler M, Lauritzen M, Rostrup E*, et al.* Early detection of Alzheimer's disease using MRI hippocampal texture. Hum Brain Mapp 2016; 37(3): 1148-61.

Szegedy C, Ioffe S, Vanhoucke V, Alemi AA. Inception-v4, Inception-ResNet and the Impact of Residual Connections on Learning. 2017; 2017.

Teng E, Chow N, Hwang KS, Thompson PM, Gylys KH, Cole GM*, et al.* Low plasma ApoE levels are associated with smaller hippocampal size in the Alzheimer's disease neuroimaging initiative cohort. Dementia and geriatric cognitive disorders 2015; 39(3-4): 154-66.

Tin Kam H. The random subspace method for constructing decision forests. IEEE Transactions on Pattern Analysis and Machine Intelligence 1998; 20(8): 832-44.

Tsao S, Gajawelli N, Zhou J, Shi J, Ye J, Wang Y, *et al.* Feature selective temporal prediction of Alzheimer's disease progression using hippocampus surface morphometry. Brain and behavior 2017; 7(7): e00733.

Winship C, Mare RD. Regression-Models with Ordinal Variables. Am Sociol Rev 1984; 49(4): 512-25.